%% file: main.tex
\documentclass[10pt,twocolumn,letterpaper]{article}

\usepackage{graphicx}
\usepackage{amsmath}
\usepackage{amssymb}
\usepackage{booktabs}

\usepackage[pagebackref,breaklinks,colorlinks]{hyperref}

\usepackage[capitalize]{cleveref}
\crefname{section}{Sec.}{Secs.}
\crefname{section}{Section}{Sections}
\crefname{table}{Table}{Tables}
\crefname{table}{Tab.}{Tabs.}

\usepackage{mytables}
\usepackage{myfigures}
\usepackage{multirow}
\usepackage{adjustbox}
\usepackage{subcaption}
\usepackage{xspace}

\usepackage{listings}
\usepackage{xcolor}
\definecolor{codegreen}{rgb}{0,0.6,0}
\definecolor{codegray}{rgb}{0.5,0.5,0.5}
\definecolor{codepurple}{rgb}{0.58,0,0.82}
\definecolor{backcolour}{rgb}{0.95,0.95,0.92}
\lstdefinestyle{mystyle}{  
    commentstyle=\color{codegreen},
    keywordstyle=\color{magenta},
    numberstyle=\tiny\color{codegray},
    stringstyle=\color{codepurple},
    basicstyle=\ttfamily\footnotesize,
    breakatwhitespace=false,         
    breaklines=true,                 
    captionpos=b,                    
    keepspaces=true,                 
    numbers=left,                    
    numbersep=5pt,                  
    showspaces=false,                
    showstringspaces=false,
    showtabs=false,                  
    tabsize=1
}
\newcommand{\update}[1]{#1}
\newcommand{\ddetr}{D\textsuperscript{2}ETR\xspace}

\newcommand{\lossl}{\mathcal{L}}
\newcommand{\blk}{\mathbf{H}}

\def\cvprPaperID{*****} % *** Enter the CVPR Paper ID here
\def\confName{CVPR}
\def\confYear{2022}

\begin{document}

\title{\ddetr: Decoder-Only DETR with Computationally Efficient Cross-Scale Attention}

\author{Junyu Lin\footnote{Works done when intern at Alibaba}\\
Nanjing University\\
{\tt\small junyulin@smail.nju.edu.cn}
\and
Xiaofeng Mao\\
 Alibaba Group\\
{\tt\small mxf164419@alibaba-inc.com}
\and
Yuefeng Chen\\
 Alibaba Group\\
{\tt\small yuefeng.chenyf@alibaba-inc.com}
\and
Lei Xu\\
 Nanjing University\\
{\tt\small xlei@nju.edu.cn}
\and
Yuan He\\
 Alibaba Group\\
{\tt\small heyuan.hy@alibaba-inc.com}
\and
Hui Xue\\
 Alibaba Group\\
{\tt\small hui.xueh@alibaba-inc.com}
}
\maketitle

%%%%%%%%% ABSTRACT
\begin{abstract}
DETR is the first fully end-to-end detector that predicts a final set of predictions without post-processing. However, it suffers from problems such as low performance and slow convergence. A series of works aim to tackle these issues in different ways, but the computational cost is yet expensive due to the sophisticated encoder-decoder architecture. To alleviate this issue, we propose a decoder-only detector called \ddetr. In the absence of encoder, the decoder directly attends to the fine-fused feature maps generated by the Transformer backbone with a novel computationally efficient cross-scale attention module. \ddetr demonstrates low computational complexity and high detection accuracy in evaluations on the COCO benchmark, outperforming DETR and its variants.
% \update{Code is available at \url{https://anonymous.4open.science/r/ddetr-7BD6}}
\end{abstract}

%%%%%%%%% BODY TEXT
\section{Introduction}

Object detection is a computer vision task to predict category labels and bounding box locations for all objects of interest in an image. Modern detectors \cite{redmon2018yolov3, tian2019fcos, he2017mask} treat this set prediction task as regression and classification problems. They optimize the predicted set with many hand-crafted components such as anchor generation, training target assignment rule, and non-maximum suppression (NMS). These components complicate the pipeline and the detectors are not end-to-end.

DETR~\cite{carion2020end-detr} proposed to build the first fully end-to-end detector, featuring an encoder-decoder Transformer architecture that predicts a final set of bounding boxes and category labels without any well-designed anchor, heuristic assignment rule, and post-processing.
The fancy design of DETR has been receiving a lot of research attention. However, DETR suffers from many problems such as slow training convergence, low performance on small objects, and high computational complexity. A bunch of works \cite{zheng2020end-act, sun2020rethinking-tsp, zhu2020deformable, yao2021efficient, meng2021conditional, gao2021fast-smca} aim to handle these critical issues. 
Many efforts have been made to efficient cross-attention to accelerate the training convergence. Multi-scale feature maps are also used to improve the accuracy of small objects. Though the works above have made some progress, the problem of high computational complexity is left untouched. 

In the field of Natural Language Processing (NLP), the OpenAI GPT series~\cite{radford2019language-gpt2,brown2020language-gpt3} adopt a decoder-only Transformer but show impressive ability on text generation. It implies that a rigorous encoder-decoder is not necessary in language modeling, which motivates us to rethink the DETR architecture. 
In DETR, the Transformer decoder is the key for locating with object queries, while the encoder, which produces self-attentive features, only plays as an assistant for the subsequent decoder. To validate it, we re-examined the overall impact of encoder by training Deformable DETR w/o encoder. We found that the encoder module brings 4.9 (+11\%) AP improvement but costs very large portion of computation, about 85 (+49\%) GFLOPs. Such a low computional efficiency of encoder prompts us to explore the possibility of removing the Transformer encoder by integrating feature extraction and self-attention based fusion functions within a single backbone, creating a simpler decoder-only detection pipeline.

\figAttnPre

In this paper, we propose {\it Decoder-only DEtection TRansformer} (\ddetr), which achieves high performance, and low computational cost with an easy architecture. To introduce the interaction among features at different locations and scales, our method leverages the advantage of Transformer backbones that providing a global receptive field of intra-scale, plus a novel module {\it Computationally Efficient Cross-scale Attention} (CECA) that performing sparse feature interaction of cross-scale via attention mechanism.
By organizing the cross-attention in which high-level feature maps as query and the remained lower-level feature as key-value pairs, CECA captures the low-level visual feature helpful for fine-grained location, but prevents the computation explosion when locating on low-level feature maps directly.
With such a design, the decoder can directly attend to the fine-fused feature maps generated by our backbone, without the aid of an encoder or other fusion block to introduce more feature interactions.

The CECA module can easily replace the encoder and go with any type of decoder to form an end-to-end detector flexibly. We cooperate it with a decoder of standard multi-head attention and a decoder of deformable attention, resulting in a vanilla \ddetr and a Deformable \ddetr.
Furthermore, we introduce two auxiliary losses: (i)  {\it token labeling loss}, which helps improve the accuracy by enhancing feature expression capabilities. (ii) \update{{\it location-aware loss}}, helps improve the localization accuracy by adding constrains the predicted bounding boxes.

\cref{fig:attnpreview} visualizes the cross-attention on different scales. Deformable DETR focuses more on low-level scales of high resolution, which requires carefully fine-fused and thus computationally inefficient. Our method focuses more on the high-level scales, which have aggregated information from all previous scales in a coarse-to-fine manner.
In evaluation on the COCO 2017 detection benchmark \cite{lin2014microsoft-coco}, our proposed methods deliver competitive performance at low computational complexity.
% e.g., 43.2 AP and 82 GFLOPs for \ddetr, 50.0 AP and 93 GFLOPs for Deformable \ddetr. 
%\update{The vanilla \ddetr achieves 43.2 AP and 82 GFLOPs at 50 epochs. The Deformable \ddetr achieves 50.0 AP and 93 GFLOPs at 50 epochs.}

\section{Related Work}

{\bf End-to-end Object Detections.}
Conventional one-stage \cite{lin2017focal-retinanet, redmon2018yolov3, tian2019fcos} and two-stage detectors \cite{ren2015faster, he2017mask, cai2018cascade} rely on anchor boxes or anchor centers.
Due to the dense anchors, an intersection-over-unit (IoU) based heuristic one-to-many assignment rule is usually applied to training, and a non-maximum suppression (NMS) is used to resolve duplications during inference.
Unlike aforementioned detectors, end-to-end detectors eliminate the need for post-processing by learning to resolve the duplicated predictions. 
DETR \cite{carion2020end-detr} initiatively applies an encoder-decoder Transformer architecture to a CNN backbone and builds an end-to-end detector. However, DETR has some issues such as slow training convergence, limited feature spatial resolution, and low performance on small objects.

Many variants have been proposed to address these critical issues. 
ACT \cite{zheng2020end-act} adaptively clusters similar query elements together. 
Deformable DETR \cite{zhu2020deformable} uses multi-scale feature maps to help detect small objects. It introduces the deformable attention mechanism that only focuses on a small fixed set of sampling points predicted from the feature of query elements. This modification mitigates the issues of convergence and feature spatial. 
Conditional DETR \cite{meng2021conditional} presents a conditional cross-attention mechanism. A spatial embedding is predicted from each output of the previous decoder layer, and then fed to next cross-attention to make the content query localize distinct regions.
SMCA \cite{gao2021fast-smca} conducts location-constrained object regression to accelerate the convergence by forcing co-attention responses to be high near initially estimated bounding box locations.
YOLOS \cite{fang2021you-yolos} argues that object detection tasks can be completed in a pure sequence-to-sequence manner.
Similarly, Pix2Seq \cite{chen2021pix2seq} treats object detection as a language modeling task conditioned on pixel inputs.
Our proposed \ddetr focuses on removing the whole encoder at minimum cost to simplify the pipeline and ease the high computation consumption. 

{\bf Multi-scale Feature Fusion.}
A line of works have been done on feature fusion and prove that a good spatial feature fusing scheme is necessary for conventional object detection, especially when detecting small objects. FPN \cite{lin2017feature-fpn} combines two adjacent feature maps and builds a top-down feature pyramid. PAN \cite{liu2018path-pan} adds an extra bottom-up path augmentation. NAS-FPN \cite{ghiasi2019fpn-nasfpn} utilizes neural architecture search to find an optimal feature pyramid structure.
In the field of end-to-end object detection, the encoder fuses feature maps via attention mechanism, playing a similar role as the feature pyramid network.
Sun \cite{sun2020rethinking-tsp} replaces the decoder with a detection head and directly uses the outputs of encoder for object prediction, which means the encoder is good at extract context features.
Yao \cite{yao2021efficient} conducts an experiment on the effect of different numbers of encoder layers and decoder layers and claims that detectors based on encoder-decoder architecture are more sensitive to the number of decoder layers, which implies the encoder is less efficient. 
These observations motivates us to seek a more economical way to exchange information on multi-scale feature maps. Our computational efficient cross-scale attention allows the backbone to generate fine-fused feature maps without the need for an encoder.

\section{Revisiting Encoder-decoder Architecture}

\update{DETR and its variants} are based on the encoder-decoder Transformer architecture. \cref{fig:framework}a show the three major parts of an end-to-end detector: backbone, encoder, and decoder. The backbone extracts \update{single or multiple} feature maps $x \in \mathbb{R}^{C\times H\times W}$ of the input image. Then the pixels in $x$ attend to each other with Transformer encoder layer, which consists of a Self-Attention (SA) and a Feed-Forward Network (FFN). The standard self-attention module~\cite{vaswani2017attention} computes attention weights based on the similarity between query-key pairs, and then compute a weighted sum over all key contents. We can fomulate the encoder as:
\begin{align}
\text{SA}(x_{q},x_{k},x_{v}) &= \text{softmax}\left(\frac{x_{q}W_{q}(x_{k}W_{k})^{T}}{\sqrt{C}}\right)x_{v}W_{v} \label{eq:revisit} \\
\text{FFN}(x) &= \sigma (xW_{1}+b_{1})W_{2}+b_{2} 
\end{align}
where $x_{q},x_{k},x_{v}= \text{Flatten}(x)$ in encoder, $\text{Flatten}(\cdot)$ is the operation that flatten $x$ along spatial dimensions. $\sigma(\cdot)$ is the non-linear activation. The self-attentive features are subsequently fed into decoder. The decoder has similar structures with an extra cross-attention module, which modifies $x_{q}$ as $o_{q}\in \mathbb{R}^{N\times C}$ in \cref{eq:revisit}. $o_{q}$ is the learned object queries.

We give a complexity analysis of encoder and decoder. For encoder whose input dimension is ${H\times W\times C}$, the complexity can be calculated as $\mathcal{O}(H^{2}W^{2}C+HWC^{2})$. Compared with encoder, the decoder adopts $N$ object queries for cross-attention, which has the complexity of $\mathcal{O}(NHWC+NC^{2})$. Generally, a relatively small number of object queries is sufficient for localization in end-to-end methods. While the amount of elements in feature map, which always a large resolution, is much greater than $N$. That is to say, the encoder have much larger computational complexity compared to the decoder, especially when the size of input feature map is large.

\section{Decoder-only DETR}

\figFramework

In essence, the encoder is a combination of intra-scale and cross-scale feature interaction. The self-attention mechanism of Transformer naturally introduces intra-scale interaction to separate feature maps. It motivates us to fill the missing cross-scale interaction and build a powerful Transformer backbone to generate fine-fused features, and further take over the inefficient encoder.

To perform feature interaction across feature maps of all scales, a naive design is to apply a dense connection to the model. 
Given $x_i$ as the original $i$-th feature map, $x_i^{j}$ as the $i$-th feature map after $j$ times of fusions, $\blk_i$ as the Transformer block of $i$-th stage.
At $i$-th stage, the feature map $x_i$ can be formulated as $\blk_i([x_1,x_2,...,x_{i-1}])$, where $[\cdot]$ denotes a concatenation of elements. 
Scales of each stage are linearly projected and spatial-wisely concatenated to the next stage to generate new scale, and do further cross-scale feature fusion. The fusing function is additional self-attention, denoted as $\text{SA}(x_q,x_k,x_v)$, where $x_q=x_k=x_v=[x_1^{i-1}, x_2^{i-2},...,x_i]$.
The final output of the backbone is $[x_1^{S},x_2^{S-1},...,x_S^{1}]$, given the last $S$ feature maps as the input of decoder.
This dense architecture is a combination of feature extraction and fusion.
However, it makes no difference from the original encoder because the low-level scales are of high-resolution and take part in almost all self-attention operations, leading to expensive computation and a waste of cross-attention from decoder.

\subsection{\ddetr Architecture}

Inspired by the dense connected Transformer in the previous section, we present a Computationally Efficient Cross-scale Attention (CECA) and build a Decoder-only DETR (\ddetr), illustrated in \cref{fig:framework}b. The architecture consists of two main components: a Transformer backbone and a Transformer decoder.
The backbone is the core of encoder-free. It contains two parallel streams, one for intra-scale interaction and another for cross-scale interaction.
Transformers that have linear computational complexity w.r.t. image size are preferred in object detection. By default, we borrow the idea of Pyramid Vision Transformer (PVT)~\cite{wang2021pvtv2} to build our backbone. We will show that \ddetr can cooperate with different Transformers in the ablation.
A decoder can learn to generate non-duplicated detections, which is the key that makes the detector end-to-end. Our \ddetr can equip any type of Transformer decoder without encoder.

{\bf Computationally Efficient Cross-scale Attention}
The idea of dense fusion is promising for integrating feature fusion to the backbone. As mentioned above, the main problem is the large number of query elements in self-attention. To address this, we decouple the intra-scale and cross-scale interaction and fuse the feature map in a sparse manner. In \cref{fig:framework}b, the backbone is divided into four Transformer stages that produce feature maps of different scales. The scale of the output feature maps progressively shrinks. All stages have a similar architecture which depends on the basic block of the chosen Transformer. Following the PVT implementation, the stage consists of an overlapping patch embedding and multiple successive Transformer layers built by spatial reduction self-attention and convolutional feed-forward module. All the feature maps are a global content aggregation within their own scale. 

The flow in parallel is the fusing stages. The fusing stage is designed for cross-scale feature fusion. Each desired scale enters a fusing stage as the query element (the feature map with red dashed line) and all existing fused scales are densely connected to the same fusing stage as the key elements. A modified spatial reduction operation is applied to key elements to alleviate the computational cost. The query scale can eventually aggregate visual features at spatial locations of all previous scales. The proposed CECA can be formulated as:
\begin{align}
{x}_i &= \blk_i({x}_{i-1}) \\
{x}_i^{*} &= \text{SA}(x_q,x_k,x_v) \\
x_q &= {x}_i, \quad x_k = x_v = [{x}_1^{*},{x}_2^{*},...,{x}_{i-1}^{*},{x}_i]
\end{align}
where ${x}_i^*$ stands for the the fused version of feature map ${x}_i$. Given the last $S$ feature maps as the input of decoder, the final output of CECA would be $[{x}_1^{*},{x}_2^{*},...,{x}_S^{*}]$.

\cref{fig:framework}c details one layer of $i$-th fusing stage. It is composed of three parts: a linear spatial reduction, a multi-head self-attention, and a feed-forward layer. ${x}_i$ stands for the feature map of the query scale.
$[{x}_1^*,{x}_2^*,...,{x}_{i-1}^*]$ stands for the feature maps of the key scales from preceding dense connections.
To prevent the query scale from losing its own high-level characteristics during information exchange, we project it with the proper channel number and concatenate to the key scales.
To reduce the computational cost, the key elements are fed to linear spatial reduction, which is an adaptive average pooling layer followed by $i$ separate $1\times 1$ convolutional layers and norm layers for feature maps of each scale. In the multi-head attention module, ${x}_i$ as the query interacts with $[{x}_1^*,{x}_2^*,...,{x}_{i-1}^*,{x}_i]$ as the keys to extract context information. 
The FFN we consider is the feed-forward with an additional depth-wise convolution. Same as the normal stage, this layer is repeated multiple times. Details of fusing stage can be found in \cref{app:sr}.

Given $h,w$ as the height and width of the last feature map $x_S$ at the last stage.
For the naive dense fusion, the complexity to compute self-attentionis
$\mathcal{O}(4^SShwP^2C)$
where $P$ is the adaptive pooling size. The complexity is dominated by the $S$, in other words, sizes of the low-level feature maps. Our CECA enjoys the cost-effective fusion emphasizing high-level feature maps. As a result, the complexity is reduced to $\mathcal{O}(ShwP^2C)$. More details can be found in \cref{app:complex}

By applying Transformer to the backbone, we introduce intra-scale interaction to pixels on separate feature maps. By adding extra fusing stages, we introduce cross-scale interaction to pixels on multi-scale feature maps. \ddetr structure fusing stages with single query scale and densely connected key elements. In this way, the number of query elements is greatly reduced, which allows us to use deeper fusing stages. Meanwhile, the subtle visual feature of low-level scales is kept to the Transformer decoder which helps to refine predictions, especially for small objects.

{\bf Vs. Transformer Encoder.} 
In end-to-end object detectors, the Transformer encoder servers as an independent feature refiner to fuse features {\it after} the backbone finished generating the very last feature map. Spatial locations at feature maps of all scales are exchanging information as equal individuals. In contrast, our proposed CECA is affiliated to the backbone, which allows the backbone to generate fine-fused feature maps. The feature interaction is split into intra-scale and cross-scale, providing by Transformer stages and fusing stages, respectively. Pixels of different scales are not symmetrical. 
Note that our backbone works as a whole so it is possible to get a better initialization in downstream tasks if CECA joins the pre-train. Moreover, because the depth of fusing stages is usually less than Transformer stages, we can further execute $(i+1)$-th stage and $i$-th fusing stage simultaneously to hide the latency introduced by the auxiliary structure. These optimizations are impossible to the conventional encoder-decoder architecture.

\subsection{Loss Function.}

Our proposed CECA backbone can generate fine-fused features and go with any type of decoder to form an end-to-end detector.
The vanilla DETR decoder applies the standard multi-head attention and uses only one feature map due to the high computational cost.
The Deformable DETR decoder employs deformable attention to extract context information from the multi-scale feature maps.
To validate the flexibility, we cooperate with the above two decoders and build \ddetr and Deformable \ddetr, which use single-scale and multi-scale feature maps, respectively.  
Futhermore, we introduce two auxiliary losses, token labeling loss and \update{location-aware} loss, the total loss is:
\begin{align}
\lossl_{total}  &=   \lossl_{cls} +  \lossl_{bbox} +  \lossl_{awr} +  \lossl_{token}
\end{align}
where $\lossl_{cls}$ denotes the classification loss, $\lossl_{bbox}$ denotes the regression loss, $\lossl_{awr}$ is the loss of aware branch, $\lossl_{token}$ is the loss of token labeling.

\figLocAware

\tabvsend

\update{{\bf Location-aware.} Predicted bounding boxes far away from corresponding target objects tend to be of low quality. 
\cref{fig:locaware} plots 10k detections' classification confidence and $IoU_{eval}$, i.e., the maximum IoU with targets of the same class. Compared with traditional detectors (Fig. 4a in \cite{wu2020iou}), we found that end-to-end detectors predict more low-quality bounding boxes that have high localization accuracy but low classification score.
To alleviate the mismatch between localization quality and detection confidence, we adopt the IoU branch~\cite{wu2020iou} and centerness branch~\cite{tian2019fcos}.
Specifically, two new branches are added to the top of all decoder layers to predict the IoU between the predicted bounding box, and predict the normalized distance from the anchor center to the target center, respectively. In practice, the reference point will serve as an anchor center. During inference, they are integrated with the classification score to suppress low-quality predictions. We formulate the aware loss term as:}
\begin{align}
\lossl_{awr} &= \frac{1}{B} \sum_{i=1}^{B} \left( \text{BCE}(\text{FFN}(\hat{y}_i), \text{IoU}(b_i,\hat{b}_i)) \right. \nonumber \\
& \qquad \qquad \left. + \text{BCE}(\text{FFN}(\hat{y}_i), \text{CTR}(b_i,\hat{b}_i)) \right)
\end{align}
\update{where $B$ is the number of bounding boxes, $\text{CTR}$ is the centerness measurement. $\hat{y}_i$ is the output corresponding to $i$-th object query, and passes through FFNs and yields predicted IoU and centerness. $b_i,\hat{b}_i$ represents the target and predicted bounding box, respectively. The higher predicted IoU and centerness value, the higher possibility for the corresponding bounding box to capture a real target. During inference, the output of classification branch is multiplied by the two branches with weighting factors to filter low-quality results. See \cref{app:aware} for more details.}

{\bf Token Labeling.}
{\it Token labeling} \cite{jiang2021all} is a novel training objective for Transformers that performs patch-wise classifications in image classification task. We adopt token labeling to train our model. 
\update{Instead of applying a model pre-trained with token labeling to the downstream object detection task, we directly introduce token labeling to the refined features at the training phase of detector.}
Specifically, we utilize mask annotations to supervise and interpolate them to align with the resolution of feature map. Each pixel is assigned with a soft token label and performs multiclass classification. %, which encourages the backbone to extract more strength features. 
\update{Note that token labeling expects a global receptive field to better classify each image patch, therefore it is suitable for vision Transformer backbones.}
\update{We found the location-specific supervision that indicates the existence of the target objects inside the corresponding local region helps not only visual grounding, classification, but also object detection. The dense supervision encourages the vision Transformer backbone to extract more strength features and facilitates localization and classification in decoder.}
The loss term of token labeling can be defined as:
\begin{align}
\lossl_{token} &= \frac{1}{B} \sum_{i=1}^{B} \sum_{j=1}^N \sum_{p,q} \text{Focal}\left(\text{FFN}(x_j[p,q]), t_j[p,q]\right) \\
t_j &= \text{Interpolate}(M, \text{SizeOf}(x_j))
\end{align}
where $x_j[p,q]$ denotes the features at position $(p,q)$ of $j$-th feature map from the backbone, $t_j$ is the corresponding target soft token label, generated by interpolating the 0-1 mask annotation $M$ to the same size of $x_j$. 
See \cref{app:token} for more details.

\section{Experiments}

{\bf Dataset.} We perform experiments on the COCO 2017 detection dataset \cite{lin2014microsoft-coco}. We train our models on the training set and evaluate on the validation set.

{\bf Implementation Details.} We follow the implementation of PVT-v2-B2-linear \cite{wang2021pvtv2}. Model pre-trained in ImageNet-1K \cite{deng2009imagenet} is used to initialize the Transformer stages of our backbone. All the fusing stages have a depth of 3, a channel number of 256, an expansion ratio of 4. Other hyperparameters are kept the same with its nearest normal stage. 
We follow the Deformable DETR training settings. The decoder has 6 layers, and the number of object queries is 300. 
For \ddetr, we use a single-scale feature map of stride 32. 
For Deformable \ddetr, two-stage mode are enabled and we use multi-scale feature maps of strides 8, 16, 32, 64, where the last feature map is obtained via a $3\times 3$ convolution layer.
The optimizer is an AdamW \cite{loshchilov2017fixing-adamw} with base learning rate of $2\times 10^{-4}$, $\beta_1=0.9$, $\beta_2=0.999$, and weight decay of $1\times 10^{-4}$. By default, we train our model for 50 epochs with a batch size of 32, and the learning rate is dropped by a factor of 0.1 at epoch 40. Random crop and resize are used for data augmentation with the largest side length set as 1333. Training and evaluation run on NVIDIA 2080Ti.

\tabvsdetrnew

\subsection{Results}

{\bf Main Results.} As shown in \cref{tab:vsend}, our method is compared with modern end-to-end detectors. We report GFLOPs for first 100 images of the validation set.
We can observe that DETR requires 500 training epochs to converge and the performance is relatively low. 
Conditional DETR boosts the convergence but the accuracy increment is less than obvious. 
Variants such as Deformable DETR, Efficient DETR, and SMCA reach better performances, yet the computational complexity remains at a high level.
The proposed \ddetr achieves a competitive 43.2 AP and the low computational complexity of 82 GFLOPs with $10\times$ fewer training epochs w.r.t. original DETR. Detailed convergence curves can be found in \cref{fig:convergence}, we run 3 different training epochs 50, 75, 108 with learning rate drops at epoch 40, 60, 80, respectively.
\update{Our method converges faster than DETR with simple backbone changing, implying the benefit of aggregating low-level features in a coarse-to-fine manner. More improvement breakdowns could be found in the ablation.}
The proposed Deformable \ddetr surpasses all end-to-end detector baselines with the highest detection accuracy of 50.0 AP and a quite low computational complexity of 93 GFLOPs.

{\bf Effect of Backbone.} 
\cref{tab:vsdetrnew} compares the proposed methods with DETR and Deformable DETR after the same training epochs of 50.
Transformer is not only a better feature extractor but also an indispensable instrument for us to conduct feature fusion in separate scales.
To understand the influence of Transformer backbone, we try DETR and Deformable DETR with PVT2. It greatly improves the performance, but the computational cost doesn't change too much. Take Deformable DETR-PVT2 as an example, it still demands a lot of effort (163 GFLOPs) on the encoder to refine the low-level feature maps. In contrast, the proposed Deformable \ddetr-PVT2 make use of high-level feature maps during cross-scale feature exchange, which delivering noticeable lower computational cost (93 GFLOPs). 
We also try to cooperate \ddetr with the CNN backbone ResNet50. It enjoys the benefits of cross-scale attention and improves the DETR-R50 baseline with 3.9 AP, yet lacking adequate intra-scale attention and the performance is 4.4 AP lower than the proposed method. This reveals the necessity of a Transformer backbone.

{\bf Effect of Encoder.}
Unlike other works using distinct attention schemes, we focus on removing the whole encoder at minimum cost to simplify the pipeline. To analyze the impact of the absence of encoder, we run DETR and Deformable DETR in the decoder-only mode, as shown in \cref{tab:vsdetrnew}.
We can see that the absence of encoder does alleviate the model complexity (up to 85 GFLOPs reduction), yet leading to a huge slope of the accuracy as well (up to 5.9 AP reduction). 
Our proposed approach prevents degradation with the computational efficient cross-scale attention and achieves high detection accuracy. 

\figConvergence

\tabvsbackbone

{\bf Compare with Other Detection Frameworks.}
We study the proposed method compared with other detection frameworks, cooperating with the popular Swin-Transformer~\cite{liu2021swin}. \cref{tab:vsbackbone} shows the result of state-of-the-art one-stage and two-stage detectors that use a Swin-Tiny backbone. We construct a Deformable \ddetr-SwinT by following the implementation of Swin-Tiny to build our normal stages. One can observe that the one-stage and two-stage detectors either consume too much computation or have too many parameters. Our method achieves comparable or even higher performance with fewer computationals and comparable parameters, demonstrating a decent generalization ability.

\update{
{\bf Inference Efficiency.}
To analyze the inference efficiency, we compare with standard DETR and Deformable DETR, as shown in \cref{tab:fps}. We use an input of batch size 1 and size $1280\times 800$ to measure the GFLOPs, time, and memory. We calculate the throughput by maximizing batch size and filling up a 16GB V100 GPU.
DETR has the cheapest time cost because it just takes the last feature map as input, which results in quite high throughput but very low performance. 
Deformable DETR benefits from the multi-scale input and gets good detection accuracy. However, it significantly increases the latency and memory cost which leads to bad throughput, even though the deformable attention is accelerated by CUDA.
On the contrary, our method has a simpler pipeline. Deformable \ddetr reduces the memory usage from 1109MB to 776MB and gets similar inference time, which nearly double the throughput.
\ddetr introduces unobvious memory cost, yet the throughput dropped mainly due to the increased latency. Considering the improved performance, \ddetr still achieves a better trade-off of accuracy and speed.
To further reduce the gap of inference time and reflect actual deployment, we optimize the detectors with TensorRT FP16 mode. The time cost of our two methods reduce from 71ms to 32ms, from 88ms to 43ms, respectively. The increased latencies compared to corresponding baselines are also mitigated, implying potential advance in throughput.
}

\tabfps

\subsection{Ablations}

\tabablation

We report the ablation study for the design choices in \cref{tab:ablation}. The decoder-only architecture brings a notable decrement to the computational complexity. The backbone changing to PVT improves the performance, but cannot completely compensate for the absence of encoder. The proposed CECA scheme increases the detection accuracy at very little cost.
\update{Note that CECA brings more accuracy gain to DETR (+7.8 AP) than Deformable DETR (+1.8 AP) because the latter already uses multi-scale feature maps. For DETR, the exploitation of low-level features grows out of nothing.}
With the auxiliary \update{location-aware} loss, we improve the performance by enhancing localization quality. Because the object queries in the decoder are randomly initialized and irrelevant to the output of backbone, adding token labeling loss will not improve the performance of \ddetr. Deformable \ddetr with two-stage mode activated uses a set of optimal features from backbone to initialize object queries and hence can make use of the token labeling enhanced features. With token labeling, Deformable \ddetr earns some free accuracy gains.

\section{Conclusion}

To simplify the pipeline of end-to-end detectors at minimal cost, we present \ddetr, which is efficient while maintaining high detection accuracy. The core is to build a Transformer that provides both internal and cross-scale feature interaction. The output feature maps from the backbone are fine-fused, thus eliminates the need for independent feature fusing phases. Our method decreases computation consumption significantly and outperforms the original DETR and its variants. We hope that our work will inspire the exploration of the potential of backbone in object detection. %In the future, we will investigate the neural architecture search of feature fusing stages.

%%%%%%%%% REFERENCES
{\small
\bibliographystyle{ieee_fullname}
\bibliography{egbib}
}

\clearpage
\input{myappendix}

\end{document}

%% file: myappendix.tex
\appendix
\section{Appendix}

\subsection{Spatial Reduction in Fusing Stage}\label{app:sr}

Given ${x}_i$ as the feature map of the query scale, $[{x}_1^*,{x}_2^*,...,{x}_{i-1}^*]$ as the preceding feature maps of the key scales. One layer of $i$-th fusing stage can be formulated as follows:
\begin{align}
x_i^* &= \text{FFN}(\text{Norm}(A)) + A \\
A &= \text{SA}(x_i, \text{SR}(\{x_k^*\}_{k=1}^{i-1}\cup \{x_i\}), \nonumber \\ 
& \qquad \qquad \ \text{SR}(\{x_k^*\}_{k=1}^{i-1}\cup \{x_i\}) ) + x_i \\
\text{SR}(\{x_j\}) &= \text{Concat}\left(\{\text{GELU}(\text{Norm}_j(\text{Pool}(x_j) W_{j}))\}\right)
\end{align}

where $\text{SR}(\cdot)$ refers to the spatial reduction operation which utilizes an adaptive poolling layer and enjoys linear computational and memory costs. $W_j \in \mathbb{R}^{C \times C} $ are learnable weights of $1\times 1$ convolutional layer. $\text{Norm}_j$ refers to layer normalization.

\subsection{Complexity for CECA}\label{app:complex}

In Pyramid Vision Transformer, the linear spatial reduction attention applys an adaptive pool of a fixed size $P$ to the key elements to reduce computational cost. The complexity is $\mathcal{O}(HWP^2C)$. 
For a multi-scale detector that using $S$ feature maps, suppose the downsampling stride is 2 and $h,w$ are the height and width of the last feature map $x_S$. The complexity becomes $\mathcal{O}((hw+4hw+...+4^{(S-1)}hw)SP^2C)$ so approximately we have $\mathcal{O}(4^{S}ShwP^2C)$. As for the CECA, only the latest scale serves as the query elements so the complexity is of $\mathcal{O}(ShwP^2C)$.

\subsection{Location-aware}\label{app:aware}

\figLocAwareAppendix

In evaluation, the IoUs between each predicted bounding box and each ground-truth target of the same class are calculated and the maximum IoU is defined as $IoU_{eval}$. \cref{fig:locawareapp} plots 10k detections of Deformable \ddetr with and without location-aware. One can see that the location-aware mitigates the mismatch between localization and classification by slightly increasing the confidence of detections with high $IoU_{eval}$.

\update{Given the coordinate of reference point $(x_r, y_r)$ and the target bounding box $(x_1^*, y_1^*, x_2^*, y_2^*)$, the target centerness is defined as:}
\begin{align}
CTR = &\sqrt{ \frac{\text{min}(x_r-x_1^*, x_2^*-x_r) \text{min}(y_r-y_1^*, y_2^*-y_r)}{\text{max}(x_r-x_1^*, x_2^*-x_r) \text{max}(y_r-y_1^*, y_2^*-y_r)} } \nonumber \\
& \times\mathbb{I}\left(x_r\in [x_1^*,x_2^*] \land y_r\in [y_1^*,y_2^*]\right)
\end{align}

Location-aware means to align the predicted bounding box with the ground-truth. We use parameters $\alpha, \beta$ to control its contribution to the classification score. The final classification score is defined as:
\begin{equation}
Score=CLS^{(1-\alpha-\beta)} \times IoU^\alpha \times CTR^\beta
\end{equation}

In the training phase, aware branches are added to the output of the backbone, the intermediate output, and the final output of decoder layers. In the inference phase, all the awareness branches are removed except the one on the last decoder layer. \update{By default, we set $\alpha=0.45$ and $\beta=0.05$ for Deformable \ddetr. Note that reference points of the vanilla \ddetr are always zero since it predicts absolute position rather than relative position, so we disable the centerness part by setting $\beta=0$.}

\figToken

\subsection{Token Labeling}\label{app:token}

% We use the mask annotation of COCO 2017 training set as supervision. Given an input image, we initialize a 3-d tensor with the same resolution and each channel represents a 0-1 mask of a category. The soft token labels are generated by downsampling this tensor to match the size of multi-scale feature maps using bilinear interpolation. We then assign the soft token label to pixels of all feature maps and add a simple FFN to perform multiclass classification. The loss functions we considered are a focal loss and a dice loss. This technique can provide each patch token the local details that help the backbone to extract better features to more accurately recognize and locate objects.
% Note that we always combine token labeling with the two-stage scheme, otherwise the object queries in the decoder are randomly initialized and never benefit from the advanced features.

With token labeling, each patch token is associated with a location-specific supervision indicating the presence of objects in the corresponding image area. It improves the object localization and recognition capabilities of vision Transformers without computational cost. The token labeling head will be excluded during inference. \cref{fig:token} visualizes the soft token label and predictions from our detector trained with token labeling. Observe that the extracted feature maps are capable of token-level classification.

\subsection{More Visualization on Multi-scales}\label{app:attn}

To better demonstrate which scale the decoder attends to, we visualize the cross-attention of the decoder on different scales of Deformable DETR and Deformable \ddetr, as shown in \cref{fig:attn}. The Transformer decoder has 6 decoder layers with one output per layer, we sum them up for readability.

We also plot the average attention on different scales for different object sizes, as shown in \cref{fig:attnsizescale}. We use the first 100 images of the validation set. Following the definition of COCO 2017 benchmark, objects whose pixel area is lower than $32\times 32$, between $32\times 32$ and $96\times 96$, larger than $96\times 96$ are defined as small, medium, and large object, respectively. Notice that the Deformable DETR pays more attention to the low-level scales and cares less about the high-level scales, especially for the small and medium objects. Instead, the proposed Deformable \ddetr can make use of the high-level scale, even when detecting small objects.

\figAttn

\figAttnSizeScale

\newpage

\begin{lstlisting}[language=Python, float=*, floatplacement=h, caption={Simplified PyTorch code of CECA based on PVTv2, details such as initialization and reshaping are omitted for readability. The entire code will be made available.}]
import torch
import torch.nn as nn

class PVTv2_CECA(nn.Module):
    def forward(self, x):
        outs = []
        for i in range(self.num_stages):
            # normal stage
            x = self.patch_embed[i](x)
            for blk in self.block[i]:
                x = blk(x)
            x = self.norm[i](x)
            # fusing stage
            if i >= self.fuse_start_lvl:
                outs.append(self.fuse_proj[len(outs)](x))
                outs_q = outs[-1:]
                outs_k = outs[:-1]
                for blk in self.fuse_block[len(outs)]:
                    outs_q = blk(outs_q, extra_key=outs_k)
                outs_q = self.fuse_norm[len(outs)](outs_q)
                outs = outs_q + outs_k
        return outs

class Block(nn.Module):
    def forward(self, x, extra_key=None):
        x = x + self.drop_path(self.attn(self.norm1(x), x + extra_key))
        x = x + self.drop_path(self.mlp(self.norm2(x)))
        return x

class Attention(nn.Module):
    def forward(self, query, key):
        Q = torch.cat([self.q(x) for x in query])
        KV = [self.kv(self.gelu(self.norm[i](self.conv1x1[i](self.pool(x))))) for i, x in enumerate(key)]
        K = torch.cat([x[0] for x in KV])
        V = torch.cat([x[1] for x in KV])
        attn = ((Q @ K) * self.scale).softmax()
        X = self.proj((attn @ V))
        return X
\end{lstlisting}

\begin{table*}[h]
\centering
\caption{Notations in the paper.}
% \begin{adjustbox}{width=0.8\textwidth}
\begin{tabular}{ll}
\toprule
Notation & Description \\
\midrule
$C$ & Channel number of the input feature map \\
$H$ & Height of the input feature map \\
$W$ & Width of the input feature map \\
$x$ & Input feature map \\
$x_q$ & Input feature map as query \\
$x_k$ & Input feature map as key \\
$x_v$ & Input feature map as value \\
$x_i^j$ & $i$-th input feature map after $j$ times of fusions \\
$x_i^*$ & $i$-th input feature map after fine-fused \\
$o_q$ & Object query \\
$\sigma$ & Activation function \\
$\blk_i$ & Transformer block of $i$-th stage \\
$W_q$ & Projection matrix for query \\
$W_k$ & Projection matrix for key \\
$W_v$ & Projection matrix for value \\
$P$ & Adaptive pooling size of spatial reduction \\
$N$ & Number of object queries \\
$S$ & Number of feature map fed into the decoder \\
$B$ & Number of bounding boxes \\
$M$ & 0-1 mask of segmentation annotation \\
$\hat{y}_i$ & Output of the $i$-th object query \\
$\hat{b}_i$ & $i$-th predicted bounding box \\
$b_i$ & $i$-th target bounding box \\
$x_i[p,q]$ & Position $(p,q)$ on $i$-th feature map \\
$t_i[p,q]$ & Position $(p,q)$ on $i$-th soft token label \\
\bottomrule
\end{tabular}
% \end{adjustbox}
\end{table*}